# AADNet: Exploring EEG Spatiotemporal Information for Fast and Accurate Orientation and Timbre Detection of Auditory Attention Based on A Cue-Masked Paradigm

Keren Shi, Xu Liu, Xue Yuan, Haijie Shang, Ruiting Dai, Hanbin Wang, Yunfa Fu, Ning Jiang, *Senior Member, IEEE,* Jiayuan He， *Member, IEEE*

**Abstract**—**Auditory attention decoding from electroencephalogram (EEG) could infer to which source the user is attending in noisy environments. Decoding algorithms and experimental paradigm designs are crucial for the development of technology in practical applications. To simulate real-world scenarios, this study proposed a cue-masked auditory attention paradigm to avoid information leakage before the experiment. To obtain high decoding accuracy with low latency, an end-to-end deep learning model, AADNet, was proposed to exploit the spatiotemporal information from the short time window of EEG signals. The results showed that with a 0.5-second EEG window, AADNet achieved an average accuracy of 93.46% and 91.09% in decoding auditory orientation attention (OA) and timbre attention (TA), respectively. It significantly outperformed five previous methods and did not need the knowledge of the original audio source. This work demonstrated that it was possible to detect the orientation and timbre of auditory attention from EEG signals fast and accurately. The results are promising for the real-time multi-property auditory attention decoding, facilitating the application of the neuro-steered hearing aids and other assistive listening devices.**

This work was supported by the Fundamental Research Funds for the Central Universities under Grant YJ202373, Science and Technology Major Project of Tibetan Autonomous Region of China under Grant XZ202201ZD0001G, 1.3.5 project for Disciplines of 1435 Excellence Grant from West China Hospital under Grant ZYYC22001, and key project from Med-X Center for Manufacturing under Grant 0040206107007. (E-mail of corresponding author: fyf@ynu.edu.cn, jiayuan.he@wchscu.cn). Yunfa Fu and Jiayuan He are co-corresponding authors. All data and images used in this article have been authorized.

Keren Shi is with Faculty of Information Engineering and Automation, Kunming University of Science and Technology, Kunming, Yunnan 650500, China. Keren Shi, Haijie Shang, Ning Jiang, and Jiayuan He are with National Clinical Research Center for Geriatrics, West China Hospital, Sichuan University, Chengdu, Sichuan 610017, China, and also with Med-X Center for Manufacturing, Sichuan University, Chengdu, Sichuan 610017, China. Xu Liu, Xue yuan and Haijie Shang are with West China Hospital, Sichuan University, Chengdu, Sichuan 610017, China. Ruiting Dai is with School of Information and Software Engineering, University of Electronic Science and Technology of China, Chengdu, Sichuan 610016, China. Hanbin Wang is with Institute of Electronic Engineering, China Academy of Engineering Physics (CAEP), Mianyang 621900, China. Yunfa Fu is with Faculty of Information Engineering and Automation, Kunming University of Science and Technology, Kunming, Yunnan 650500, China.

## I. INTRODUCTION

HUMANS have the ability to concentrate on the voice of a particular speaker in a noisy environment, known as "Cocktail Party Effect" [1]. It is attributed to the ability of the brain filtering out irrelevant sounds and selectively processing the interested content received. Electroencephalogram (EEG) is the electrical manifestation of brain activities and a measurement of the cortical activities. It was demonstrated that the amplitude envelope of the attended speech was represented in the oscillatory activities in the human cortex [2], [3], [4], [5], [6], [7]. The findings provided the evidence for the possibility of decoding the selective attention of the auditory from the measurement of the cortical activities, including in an invasive [8] and noninvasive way [9]. EEG is the electrical manifestation of brain activities. Considering its non-invasive and low-cost properties, EEG-based auditory attention decoding (AAD) could be potentially applied in hearing aids and other assistive listening devices, achieving the effortless control through brain signals and improving speech comprehension in noisy and loud settings [10].

Many methods have been proposed for EEG-based AAD. They could be mainly categorized into two groups, i.e., audio stimulus reconstruction and neural recording classification. The first was designed to reconstruct the audio stimulation from EEG signals and perform the correlation analysis between the original and constructed stimulus representations. It needed the information of original audio source in generating the results and the performance was not good with short decision windows [11], [12], [13]. The second generally employed machine learning (ML) and deep learning (DL) methods for directly classifying EEG signals. It did not need the information of original audio source in model testing, and its performance with short decision windows was better than that of the first [14], [15], [16], [17], [18], which made it suitable for real-world applications. This study focuses on the second method with scenarios of short decision windows.

Auditory spatial attention detection (ASAD) is a common type of AAD [19]. It was designed for discerning the spatial focus of the attention regardless of the audio content, which was suitable for the method of neural recording classification. Most previous ASAD studies rely on the public datasets, such as KUL (16 subjects) [20], DTU (18 subjects) [21], which employed an experimental protocol that informed the participants in advance to focus on sounds coming from a specific direction. In these circumstances, the participants would be prepared and waited for the sound to come, instead of selecting the sound that they are interested in and distinguishing it from the other sound sources in the real-world scenario. The difference in the timing of paying attention might induce changes in EEG signals. As such, to imitate the cocktail party scenario, the protocol needed to be adjusted to conceal direction information in the cue for the participants.

ML and DL methods were both employed for ASAD task. ML methods need feature extraction or feature engineering, i.e., calculating features manually from EEG signals, and sent to the classifiers. DL methods do not rely on manual feature extraction and could receive raw data as input and achieve an end-to-end classification [22]. Cai et al. used the β band in EEG, proposed a neural attention mechanism for EEG convolutional networks, EEG-Graph Net, which consists of three modules: a graph representation module, a biologically inspired channel-wise attention module, and a graph structure learning mechanism. It achieves an average accuracy of 96.1% and 78.7% within 1 second on the KUL and DTU databases, respectively [17]. Geirnaert et al. introduced the decoding of the directional focus of attention using filter bank common spatial pattern filters (FBCSP) as an alternative AAD paradigm, they down-sampled to 64Hz, applied bandpass filtering in different frequency bands, and utilized FBCSP filters along with LDA classifiers, achieving a high accuracy of 80% for 1-second windows and 70% for quasi-instantaneous decisions [14], [18]. Vandecappelle et al. down-sampled to 128Hz and applied bandpass filtering of EEG signals in the 1-32Hz range, presenting a convolutional neural network (CNN)-based approach that achieved a median accuracy of around 81% within 1-2 seconds [15]. Fan et al. proposed a dynamical graph self-distillation (DGSD) approach, applying bandpass filtering in the 1-50Hz range and Z-normalizing each trial. They represented the non-Euclidean EEG data as graph signals, effectively extracting key features of spatial auditory attention by combining graph convolutional network (GCN) operations with self-distillation, achieving accuracies of 90.0% and 79.6% on the KUL and DTU datasets, respectively, under a 1-second time window [16].

The length of the decision window was related to the delay of the system. The short decision window would improve the system response time, making it applicable in more areas. However, with short EEG windows, the data was limited and the algorithms needed to extract as much information as possible for accurate identification [23]. Many studies employed only spatial or temporal representations of EEG, or extracting coarse-grained information, failing to enable short-time decision-making. To maximize the use of spatiotemporal information from the limited EEG data, we propose a model based on spatiotemporal hybrid decoding, AADNet. This model includes three key modules: the Temporal Learning Module, the Spatial Learning Module, and the Hybrid Decoding Module. It balances the extraction of fine-grained temporal information and the capture of channel-related spatial information, achieving deep fusion of spatiotemporal features and improving the performance in both learning capability and decision-making efficiency.

The main contributions of this work are as follows:

A cue-masked auditory attention paradigm was proposed, and a dataset was provided. It changed the cue from sound direction to sound content, which was timbre in this study, obfuscating the directional focus of attention for better simulating complex real-world scenarios, benefiting EEG-based AAD studies.

A decoding model was proposed for EEG-based AAD task with short decision windows. It extracted spatial and temporal features simultaneously. Compared to previous methods, it significantly improved the performance of both orientation attention (OA) and timbre attention (TA) detection.

The difference in decoding OA and TA was analyzed. It improved the understanding of auditory attention processing of AAD tasks and provided insights into the decoding model design.

## II. METHOD

### A. Participants

Sixteen healthy subjects (ages 17-32, 6 females and 10 males) participated in the study. The experimental procedures were provided and written informed consent was obtained before the experiment. The experiment was in accordance with the Declaration of Helsinki and approved by the Research Ethics Committee of West China Hospital, Sichuan University (# 2024582). All data and images were obtained with written

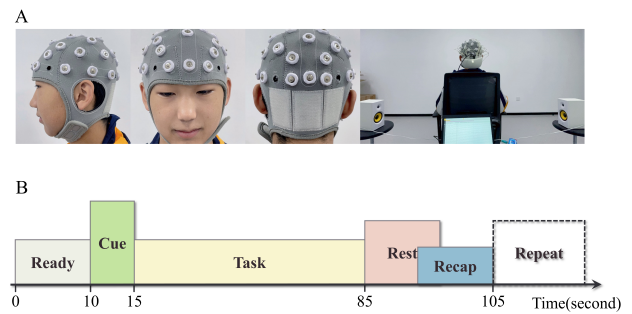

Fig. 1. Illustrations of the experimental protocol. (A) Schematic diagram of participant EEG cap placement and speaker arrangement, (B) Experimental procedure. The consent has been obtained from the participant for personal image publication. To help participants focus on the audio, after completing the task, they first entered a rest phase and, once calm, recounted the audio story from the previous trial.

consent from the participants for analysis and publication.

## B. Experimental Protocol

**TABLE I**
**AADNET ARCHITECTURE**

| Component | Layer | KernelNum | Size | Activation | Options |
|---|---|---|---|---|---|
| 1 | Conv2D | 32 | (1, 64) | Linear | |
| | BatchNorm2D | | | | |
| 2 | Conv2D | 64 | (32, 1) | Linear | |
| | BatchNorm2D | | | | |
| | Activation | | | ELU | |
| | AvgPool2D | | (1, 4) | | |
| | Dropout | | | | $p = 0.25$ |
| 3 | Conv2D | 64 | (1, 16) | | |
| | BatchNorm2D | | | | |
| | Activation | | | ELU | |
| | Conv2D | 64 | (1, 1) | | |
| | BatchNorm2D | | | | |
| | Activation | | | ELU | |
| | AvgPool2D | | (1, 8) | | |
| | Dropout | | | | $p = 0.25$ |
| Classifier | Linear | 64 | | | |
| | Linear | | | 2 | SoftMax |

**TABLE II**
**SETTINGS OF AADNET HYPER-PARAMETERS AND THE HYPER-PARAMETER SEARCH GRID**

| Hyper-parameter | Value | Grid |
|---|---|---|
| Learning rate $p$ | $10^{-3}$ | $[10^{-1}, 10^{-2}, 10^{-3}, 10^{-4}]$ |
| Batch size | 20 | $[10, 20, 50]$ |
| Epochs | 100 | $[20, 50, 100, 150, 200]$ |
| Weight decay | $10^{-2}$ | $[10^{-1}, 10^{-2}, 10^{-3}]$ |

The experiment was conducted in a soundproof room, with the field of view for the subjects restricted to white walls, as shown in Fig. 1. The subjects were exposed to mixed-gender audio stimuli. The audio materials were sourced from text-to-speech generated story texts, featuring standard male and female voices with a sampling rate of 48kHz. To simulate natural speech scenarios, the audio material pool included 24 speech files, with 12 male and 12 female voice recordings. The hybrid audio stimuli consisted of two speech segments (one male and one female), played from speakers positioned to the left and right of the subjects. The two speech segments had equal root mean square (RMS) intensity, and their timbre differences were analyzed using spectral analysis, as shown in Fig. 2. The duration of each speech stimulus was 69 seconds, defined as one trial. Overall, we generated audio from text (male and female voices), processed each segment to ensure equal volume and length, and then stored them in separate male and female audio pools for extraction.

Each participant underwent 12 trials of hybrid audio stimulus experiments, where they were tasked to perform dual-attention tasks (OA and TA target) following prompts. For example, after being prompted with an attention cue for male voices, participants were required to selectively concentrate their attention on the output content of the male timbre speech segment and locate its auditory position. Simultaneously, the output content and orientation of the female timbre speech segment were instructed to be ignored as noise interference. The targets for timbre and orientation in the hybrid audio stimulus trials were randomized, but the number of trials for different timbres and orientations were equal.

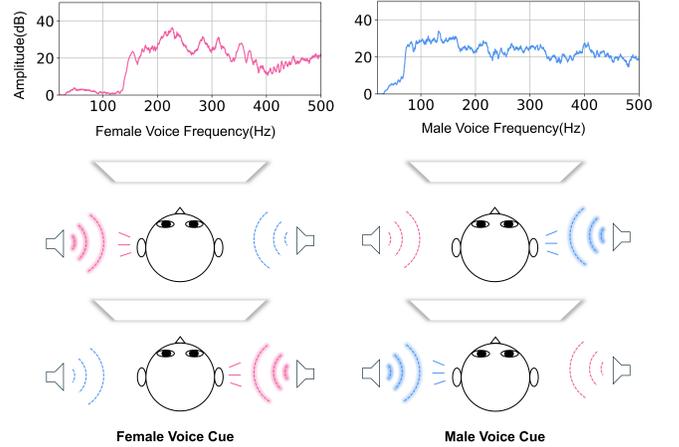

Fig. 2. The spectrum representation of audio from the same spoken content, with pink indicating female voice and blue indicating male voice in the figure. Through the cue of a female or male voice, participants are informed about the upcoming sound they need to focus on. At the onset of the task, participants are required to direct their attention immediately to the sound corresponding to the cue and maintain this focused state. One side of the audio serves as the target stimulus, while the other acts as noise, with the target appearing randomly on the left or right. Since participants cannot predict whether the sound will originate from the left or right before the task begins, they are unable to preemptively orient their orientational attention.

EEG signals were recorded simultaneously with audio cues during the experiment with a commercial wireless amplifier (Enobio EEG systems, NE Neuroelectrics, Spain). 32 electrodes were employed based on the standard 10-20 system, covering the entire brain. The signals were sampled at a frequency of 500 Hz.

## C. AADNet

An advanced DL framework, *i.e.*, AADNet, was proposed to rapidly decode participants' OA and TA. The structure is illustrated in Fig. 3. It has three modules. Initially, the temporal learning module (Block1) models the raw sequence $S \in \mathbb{R}^{B \times 1 \times C \times T}$ and passes the output feature maps $F \in \mathbb{R}^{B \times K \times C \times T}$ to the spatial learning module (Block2) to generate spatiotemporal fused features $F_{fused}$. Subsequently, the fusion decoding module (Block3) decodes the spatiotemporal features and performs dimensionality reduction, producing attentional markers $F_{target}$. Here, $B$ denotes the number of decision windows per batch, $C$ represents the number of channels, $T$ denotes the number of decision time points, and $K$ represents the number of generated temporal feature maps. Finally, the classification block uses $F_{target}$ to make attentional decisions.

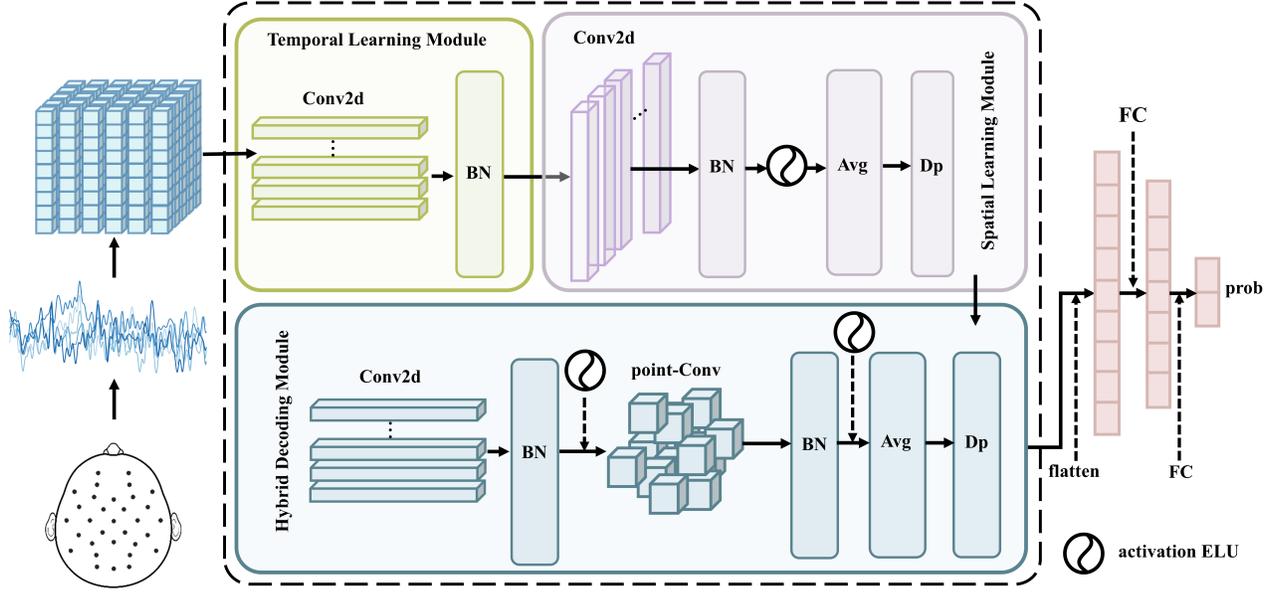

Fig. 3. Model structure. AADNet contains three modules, the Temporal Learning Module, the Spatial Learning Module, and the Hybrid Decoding Module. Avg denotes an average pooling layer, FC denotes a fully-connected layer, ELU denotes an activation layer.

The details of each module were described as follows.

### 1) Temporal Learning Module

The temporal learning module comprises $K$ temporal filters, specifically 2D convolutional kernels (where $K = 32$), each kernel having a size of $(1, 64)$. These filters produce $K$ feature maps $F$ that contain EEG signals filtered across different band-pass frequencies. Additionally, the temporal convolutional kernels have a length of 64, enabling the capture of frequency information at 4Hz and above. The temporal features $F$ for each batch of time series $S = \{s_1, s_2, \ldots, s_i, \ldots, s_B | s_i \in \mathbb{R}^{C \times T}\}$ are computed as follows:

$$F = \text{BatchNorm}(\text{Conv}(S)) \tag{1}$$

### 2) Spatial Learning Module

The spatial learning module uses depthwise convolution to model the temporal features $F$. This module, comprising $C$ channel filters, provides a direct means of learning the spatial representation for each temporal filter, effectively extracting spatial features across different temporal scales. Additionally, the use of depthwise convolution with non-fully connected feature maps reduces the number of trainable parameters. The spatiotemporal fused features $F_{fused}$ for each batch of samples are computed as follows:

$$F_{f1} = \text{BatchNorm}(\text{Conv}(F)) \tag{2}$$

$$F_{f2} = \text{Avgpooling}(\text{ELU}(F_{f1})) \tag{3}$$

$$F_{fused} = \text{Dropout}(F_{f2}) \tag{4}$$

where $F_{f1}$ and $F_{f2}$ denote intermediate computation results of $F_{fused}$. Batch normalization is applied along the feature map dimension, and dropout techniques are used to enhance the model's generalization capability. Both convolutions use ELU (Exponential Linear Unit) activation functions, as the model's performance significantly improves with nonlinear activation. An average pooling layer is employed to down-sample the decision signal to a sampling rate of 125Hz.

### 3) Hybrid Decoding Module

The hybrid decoding module employs a sequence of convolutions to model the spatiotemporal fused features $F_{fused}$. This convolutional sequence includes a depthwise separable convolution followed by a pointwise convolution. These convolutions decode the temporal and spatial information contained in $F_{fused}$ and output the attention labels $F_{target}$. The attention labels $F_{target}$ for each batch of samples are computed as follows:

$$F_1 = \text{ELU}\left(\text{BatchNorm}\left(\text{Conv}\left(F_{fused}\right)\right)\right) \tag{5}$$

$$F_2 = \text{ELU}\left(\text{BatchNorm}(\text{PointConv}(F_1))\right) \tag{6}$$

$$F_{target} = \text{Dropout}(\text{Avgpooling}(F_2)) \tag{7}$$

where $F_1$ and $F_2$ denote intermediate computation results of $F_{target}$.

### 4) Hyperparameter Settings

A grid search on a validation set was conducted to determine a set of reasonable values for hyperparameter optimization, as listed in Table I. The Adam optimizer was employed alongside weight decay techniques, dropout, and batch normalization to mitigate overfitting and enhance generalization. The detailed configuration of the AADNet is described in Table II.

ACC (%) COMPARISON OF AADNET AND FIVE DL AND ML MODELS IN ORIENTATIONAL ATTENTION DECODING AND TIMBRE ATTENTION DECODING ACROSS THREE DECISION WINDOWS

| Task | Window Length (s) | AADNet | EEGNet | ShallowCovNet | DeepCovNet | FBCSP+SVM | PCA+SVM |
|------|------|------|------|------|------|------|------|
| OA | 0.1 | **91.81 ± 3.16** | 87.39 ± 4.53 | - | 79.61 ± 9.38 | - | 55.78 ± 4.49 |
|  | 0.5 | **93.46 ± 2.97** | 88.11 ± 5.63 | 82.11 ± 7.98 | 80.89 ± 8.55 | 61.38 ± 10.15 | 55.11 ± 4.80 |
|  | 1 | **89.09 ± 4.37** | 78.52 ± 8.41 | 75.75 ± 8.59 | 75.21 ± 9.80 | 56.02 ± 18.08 | 54.50 ± 5.81 |
| TA | 0.1 | **89.87 ± 3.34** | 84.29 ± 4.33 | - | 77.17 ± 6.11 | - | 58.26 ± 5.03 |
|  | 0.5 | **91.09 ± 4.18** | 84.33 ± 5.69 | 76.85 ± 7.43 | 75.96 ± 6.71 | 55.65 ± 16.16 | 57.78 ± 5.44 |
|  | 1 | **84.06 ± 6.88** | 72.29 ± 7.59 | 69.39 ± 6.74 | 70.73 ± 6.05 | 57.37 ± 12.96 | 57.58 ± 6.38 |

SPE (%) COMPARISON OF AADNET AND FIVE DL AND ML MODELS IN ORIENTATIONAL ATTENTION DECODING AND TIMBRE ATTENTION DECODING ACROSS THREE DECISION WINDOWS

| Task | Window Length (s) | AADNet | EEGNet | ShallowCovNet | DeepCovNet | FBCSP+SVM | PCA+SVM |
|------|------|------|------|------|------|------|------|
| OA | 0.1 | **91.53 ± 5.15** | 86.13 ± 8.87 | - | 78.47 ± 17.44 | - | 48.87 ± 22.12 |
|  | 0.5 | **93.24 ± 4.48** | 87.00 ± 7.78 | 82.20 ± 11.67 | 80.95 ± 16.12 | 36.14 ± 17.57 | 47.44 ± 18.73 |
|  | 1 | **88.40 ± 7.76** | 76.43 ± 12.50 | 76.91 ± 11.72 | 76.16 ± 15.90 | 40.61 ± 23.14 | 45.49 ± 18.78 |
| TA | 0.1 | **86.40 ± 7.27** | 81.00 ± 9.68 | - | 73.88 ± 12.54 | - | 53.62 ± 32.73 |
|  | 0.5 | **88.43 ± 7.76** | 82.39 ± 9.71 | 75.72 ± 12.50 | 69.83 ± 16.14 | 39.78 ± 31.40 | 50.71 ± 34.49 |
|  | 1 | **79.84 ± 12.86** | 71.35 ± 10.36 | 67.79 ± 12.29 | 66.10 ± 16.13 | 48.98 ± 36.62 | 52.17 ± 33.87 |

### D. Performance Evaluation

Neural networks profit from broadband EEG input [24]. In our study, we applied finite impulse response (FIR) bandpass filtering in the range of 0.4-32 Hz [25]. Average referencing and independent component analysis (ICA) [26] were employed sequentially to remove artifacts such as eye movements and muscle activity [27]. The preprocessed signals were segmented into windows and taken as the input of the model. Three window lengths were evaluated, 0.1s, 0.5s and 1s. The performance of the proposed method was compared with five traditional methods of EEG classification, which were EEGNet [28], shallow convolutional neural network (ShallowCovNet) [29], deep convolutional neural network (DeepCovNet) [29], support vector machine with principal component analysis (PCA+SVM) [30], and support vector machine with filter bank common spatial pattern (FBCSP+SVM) [14]. To effectively extract EEG features at low computational costs, a lightweight convolutional neural network, EEGNet, has been proposed specifically for handling EEG data. Shallow ConvNet rapidly extracts low-level features of EEG data using a small number of convolutional and pooling layers, making it suitable for real-time applications and systems requiring quick responses. Deep ConvNet, on the other hand, is suitable for complex EEG signal classification tasks, capable of extracting hierarchical features and demonstrating excellent performance when trained on large-scale datasets. FBCSP decomposes EEG signals into frequency bands and applies CSP to extract features from different frequency bands, thereby enhancing feature discriminability. PCA helps in removing noise and redundant information, and the reduced-dimensional features are inputted into an SVM classifier for classification.

We evaluated the performance of the proposed method and all other methods on a subject-by-subject basis. For each participant, we divided the data into five non-overlapping grand folds. One grand fold was used for testing, while the remaining four grand folds were used for hyper-parameter tuning through inner 5-fold cross-validation. This process was repeated five times across the five grand folds, and the average results were calculated. The metrics, including accuracy (ACC), f1-score (F1), precision (PRE), sensitivity (SEN), and specificity (SPE), were calculated to evaluate the performance of the methods [17], [31], [32]. Specifically, ACC is the proportion of correctly predicted samples to the total number of samples. The calculation formula is as follows:

$$ACC = \frac{TP + TN}{TP + TN + FP + FN} \qquad (8)$$

$TP$ is the number of correctly predicted positive samples, $TN$ is the number of correctly predicted negative samples, $FP$ is the number of incorrectly predicted positive samples, and $FN$ is the number of incorrectly predicted negative samples. In the OA task, we choose the right side as the positive example, and in the TA task, we choose female voices as the positive example. The ROC curve is a plot of the true positive rate (SEN) against the false positive rate (1-SPE) at various threshold settings. The F1 is the harmonic mean of PRE and Recall (SEN). The calculation formula is as follows:

$$F1 = \frac{2 \cdot PRE \cdot SEN}{PRE + SEN} \qquad (9)$$

| Task | Window Length (s) | **AADNet** | EEGNet | ShallowCovNet | DeepCovNet | FBCSP+SVM | PCA+SVM |
|------|------|------|------|------|------|------|------|
| OA | 0.1 | **91.81 ± 4.75** | 88.00 ± 6.62 | - | 79.78 ± 12.27 | - | 60.84 ± 19.70 |
| | 0.5 | **93.61 ± 4.98** | 88.89 ± 6.58 | 81.39 ± 9.81 | 79.69 ± 15.07 | 59.67 ± 16.89 | 60.14 ± 17.16 |
| | 1 | **89.24 ± 7.47** | 79.89 ± 10.52 | 74.04 ± 10.88 | 73.27 ± 17.50 | 62.51 ± 17.78 | 60.05 ± 16.92 |
| TA | 0.1 | **92.84 ± 4.05** | 85.75 ± 10.25 | - | 78.22 ± 14.57 | - | 53.62 ± 34.15 |
| | 0.5 | **93.34 ± 4.83** | 85.13 ± 8.10 | 74.89 ± 17.20 | 79.99 ± 13.34 | 56.09 ± 30.83 | 54.62 ± 34.46 |
| | 1 | **87.37 ± 8.85** | 71.26 ± 13.07 | 68.70 ± 13.46 | 72.40 ± 15.96 | 51.62 ± 36.43 | 52.70 ± 33.13 |

PRE is the proportion of true positive predictions out of all positive predictions, indicates the accuracy of positive predictions. The calculation formula is as follows:

$$PRE = \frac{TP}{TP + FP} \qquad (10)$$

SEN is the proportion of true positive predictions out of all actual positives, indicates the model's ability to capture positive samples. The calculation formula is as follows:

$$SEN = \frac{TP}{TP + FN} \qquad (11)$$

SPE is the proportion of true negative predictions out of all actual negatives, indicates the accuracy of negative predictions. The calculation formula is as follows:

$$SPE = \frac{TN}{TN + FP} \qquad (12)$$

## III. Results

### A. Performance Comparison Among the Models

DL methods outperform ML methods in overall classification, indicating alongside previous research [15] that nonlinear ML approaches can aid in swiftly and reliably decoding auditory attention (orientational and timbre). Notably, ShallowConvNet offers a lightweight model architecture, DeepConvNet can capture more complex features, and EEGNet stands out for its ability to decouple spatial and temporal features, yet none match the performance of AADNet. The proposed model, AADNet, outperformed the other methods in both tasks. To comprehensively compare the performance of various models, obtain the optimal decision window length parameter, and evaluate the temporal scale sensitivity of our model, we conducted parameter comparison experiments on the proposed AADNet model across different scales. We consider EEGNet as the optimal baseline model.

#### 1) Comparison of AADNet with Different Models on OA and TA Tasks

In the OA task, we compared our proposed model with five other models at 0.1-second, 0.5-second, and 1-second decision windows. The best detection performance was achieved with a 0.5s decision window, with an average auditory attention decoding ACC of 93.46% and a standard deviation (SD) of 2.97%. Our proposed model outperformed other models at all decision window lengths, with ACC of 91.81% (SD: 3.16%) at 0.1s and 89.09% (SD: 4.37%) at 1s. In the TA decoding task, the average ACC for decision windows of 0.1s, 0.5s, and 1s were 89.87% (SD: 3.34), 91.09% (SD: 4.18%), and 84.06% (SD: 6.88%), respectively. In the OA and TA decoding task, the accuracy of our model for each subject is shown in Fig. 4.

In our study, 0.5 seconds was determined as the optimal decision window length, a choice closely related to the sample size of the study and the spatiotemporal characteristics of the task. A 0.5-second time window is sufficient to capture enough spatiotemporal information, thus providing ample EEG signal features for decoding. Especially when dealing with high-noise EEG signals, an appropriate time window can balance information extraction and noise suppression, which helps improve decoding accuracy. Moreover, the sample size in this study is large enough to ensure that representative and stable EEG data can be obtained within the 0.5-second time window, providing strong support for precise decoding.

The superior decoding accuracy of AADNet compared to the baseline model can be attributed to its design advantages. First, AADNet uses a deep neural network architecture capable of automatically extracting complex spatiotemporal features from

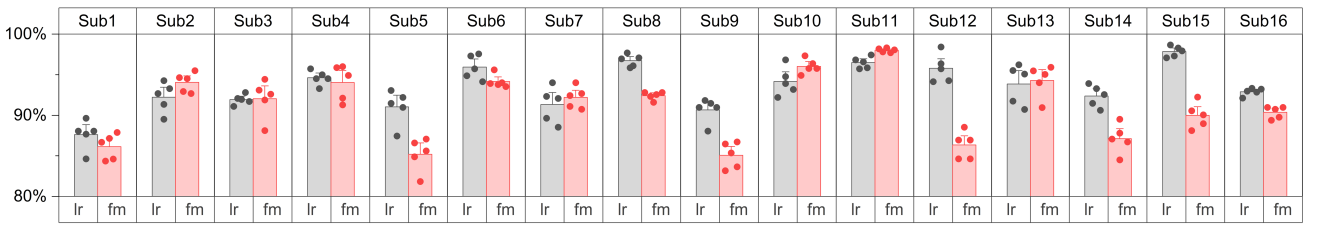

Fig. 4. Detection accuracies (%) for each subject in the OA (gray) and TA (red) tasks under 0.5-second decision window. Sort the horizontal axis by the subject IDs, OA (lr) and TA (fm) tasks.

TABLE VI

PRE (%) COMPARISON OF AADNET AND FIVE DL AND ML MODELS IN ORIENTATIONAL ATTENTION DECODING AND TIMBRE ATTENTION DECODING
ACROSS THREE DECISION WINDOWS

| Task | Window Length (s) | AADNet | EEGNet | ShallowCovNet | DeepCovNet | FBCSP+SVM | PCA+SVM |
|------|------|------|------|------|------|------|------|
| OA | 0.1 | **92.35 ± 3.98** | 87.80 ± 5.75 | - | 82.82 ± 8.92 | - | 56.57 ± 5.98 |
| | 0.5 | **93.95 ± 3.77** | 88.36 ± 6.09 | 83.92 ± 8.73 | 84.28 ± 9.57 | 49.28 ± 12.83 | 55.41 ± 5.65 |
| | 1 | **90.06 ± 5.49** | 79.33 ± 8.22 | 78.09 ± 8.84 | 78.65 ± 10.78 | 53.05 ± 14.86 | 54.62 ± 7.16 |
| TA | 0.1 | **87.12 ± 5.15** | 82.02 ± 5.56 | - | 74.55 ± 9.46 | - | 55.35 ± 9.17 |
| | 0.5 | **89.18 ± 6.02** | 82.50 ± 7.42 | 74.54 ± 11.22 | 73.02 ± 9.89 | 49.79 ± 10.27 | 54.66 ± 6.90 |
| | 1 | **82.85 ± 9.81** | 70.06 ± 11.31 | 66.88 ± 11.37 | 67.86 ± 9.77 | 52.91 ± 12.74 | 55.88 ± 7.33 |

TABLE VII

F1 (%) COMPARISON OF AADNET AND FIVE DL AND ML MODELS IN ORIENTATIONAL ATTENTION DECODING AND TIMBRE ATTENTION DECODING
ACROSS THREE DECISION WINDOWS

| Task | Window Length (s) | AADNet | EEGNet | ShallowCovNet | DeepCovNet | FBCSP+SVM | PCA+SVM |
|------|------|------|------|------|------|------|------|
| OA | 0.1 | **91.92 ± 3.24** | 87.64 ± 4.61 | - | 81.45 ± 7.48 | - | 56.54 ± 11.59 |
| | 0.5 | **93.54 ± 3.11** | 88.40 ± 5.53 | 82.46 ± 8.06 | 79.99 ± 10.45 | 56.56 ± 9.86 | 55.28 ± 10.17 |
| | 1 | **89.11 ± 4.94** | 78.97 ± 8.31 | 75.38 ± 9.05 | 73.66 ± 13.73 | 58.89 ± 13.52 | 54.17 ± 10.94 |
| TA | 0.1 | **89.70 ± 3.28** | 83.34 ± 6.41 | - | 75.94 ± 10.85 | - | 50.99 ± 23.46 |
| | 0.5 | **90.83 ± 4.12** | 83.39 ± 6.74 | 73.86 ± 13.89 | 75.21 ± 9.80 | 60.17 ± 9.50 | 59.70 ± 12.23 |
| | 1 | **83.88 ± 7.74** | 69.76 ± 11.74 | 67.04 ± 11.42 | 68.84 ± 11.36 | 64.59 ± 11.24 | 57.60 ± 12.75 |

raw EEG signals without relying on manual feature engineering. This automatic feature extraction ability allows AADNet to better adapt to the complexity of the signals and capture finer-grained spatiotemporal information, thereby improving decoding accuracy. Compared to traditional baseline models, AADNet efficiently utilizes the limited signal data within a shorter time window, extracting more useful information and avoiding data loss.

Additionally, AADNet exhibits smaller fluctuations in SD, indicating higher decoding stability across different samples and tasks. The smaller SD suggests that the model can maintain consistent high performance across various subjects and experimental conditions, avoiding overfitting or dependency on specific data. This characteristic makes AADNet more reliable in practical applications, capable of providing stable decoding results under different environmental and noise conditions.

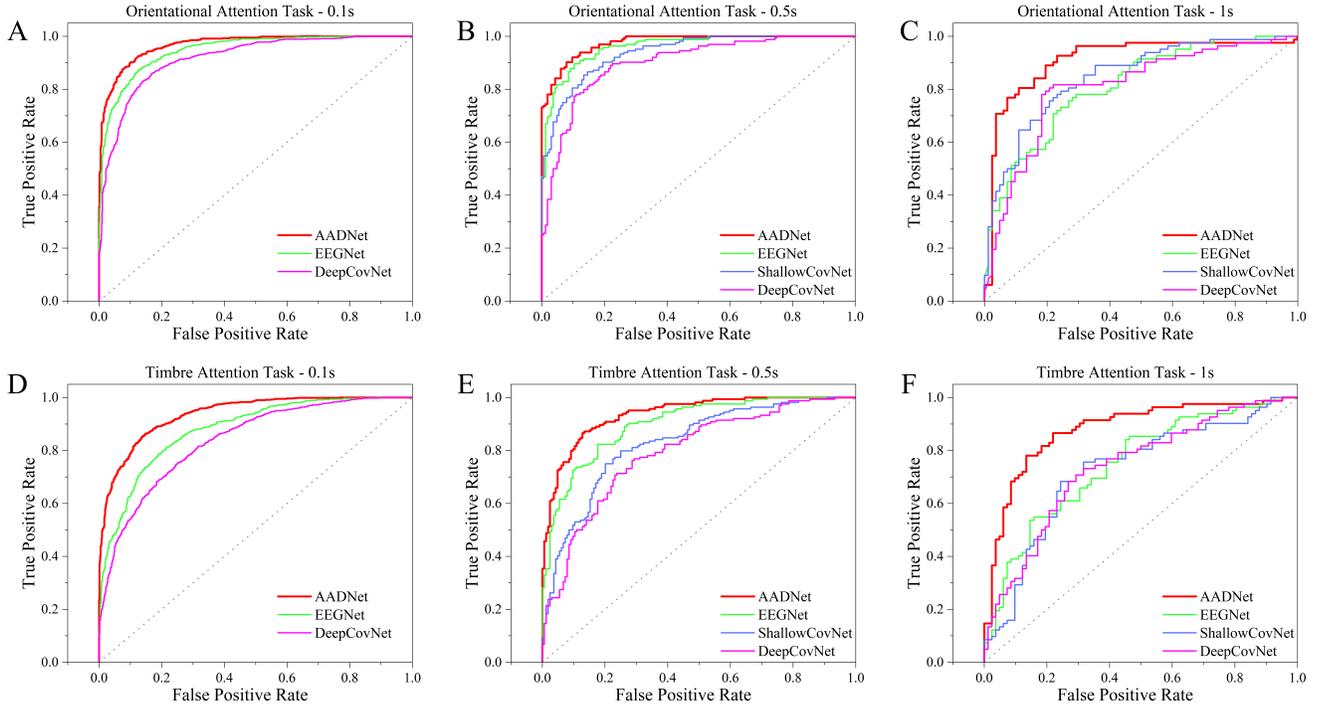

Fig. 5. Comparison of ROC performance between the proposed AADNet model and ML, with (A), (B), and (C) representing the orientational attention task for decision windows of 0.1s, 0.5s, and 1s respectively. (D), (E), and (F) represent the timbre attention task for decision windows of 0.1s, 0.5s, and 1s respectively. AADNet consistently outperforms across both short (0.1s) and longer (1s) decision windows.

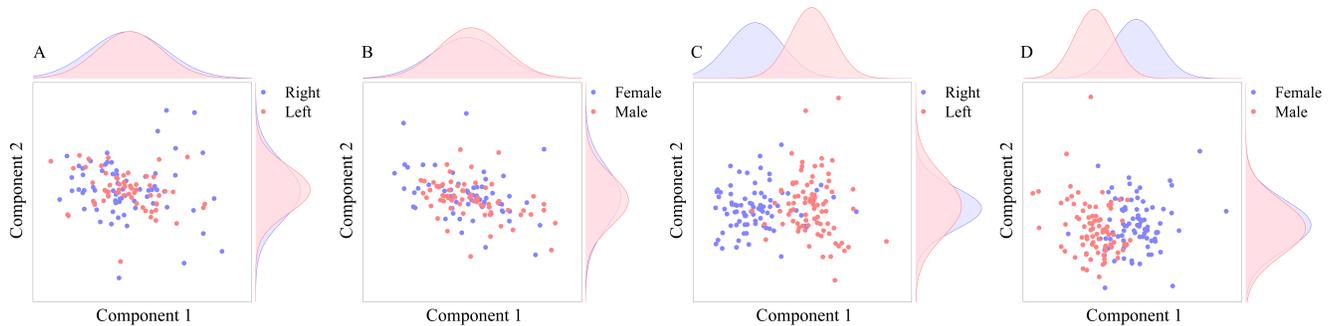

Fig. 6. Scatter plots of component comparisons. (A) Raw EEG data from orientational attention experiments, (B) Raw EEG data from timbre attention experiments, (C) Orientational attention markers visualized after extraction by AADNet, and (D) Timbre attention markers visualized after extraction by AADNet. Indistinguishable raw EEG data becomes distinguishable after processing with the AADNet.

The effective duration of the mixed speech stimulus was 70 seconds (effective stimulation time is 69 seconds). We compared the classification performance of the proposed AADNet model with traditional ML models (FBCSP+SVM, PCA+SVM) and DL models (DeepConvNet, ShallowConvNet, EEGNet) using metrics such as ACC, F1, PRE, SEN, and SPE for the dual attention decoding task, as shown in Tables III to VII. Additionally, we plotted the ROC curve for comparison [33], as shown in Fig. 5. Since the 0.1s decision window only contains 50 timepoints, ShallowConvNet and FBCSP+SVM could not effectively learn and thus are not shown for the 0.1s decision window. The classification performance of ML models was only slightly better than random classification and is therefore not displayed in the figure. Overall, the proposed AADNet model performs exceptionally well at high temporal resolution, which corresponds to the time required for human auditory attention shifts. Other auditory attention detection models have shown similar performance at such low latency settings (approximately 100 milliseconds) [34], [35], [36], [37], [38], [39], [40]. These results indicate that real-time decoding of auditory orientational and timbre attention is achievable.

### 2) Comparison of OA and TA Tasks

We found that, when using DL models for auditory attention decoding, the performance of the model in the OA task always outperforms that in the timbre attention TA task. Specifically, the decoding accuracy for OA tasks is usually higher, and the SD is generally smaller, indicating stronger stability of the DL model in the OA task. This result suggests that the features captured by the EEG signal in the OA task are more easily learned by the model compared to the TA task. We speculate that the OA task may involve more robust brain responses related to spatial localization and attention resource allocation, which could make the neural representation of the OA task more prominent and thus more conducive to high-accuracy decoding by DL models.

In contrast, EEG signals in the TA task exhibit greater complexity and variability during decoding. This could be due to the fact that the brain's processing of sound in the timbre attention task relies more on subtle changes in audio features, which may be more dispersed or complex in the EEG signals. As a result, DL models do not perform as strongly in the TA task, and in some cases, the decoding accuracy and stability are lower.

In comparison, traditional ML models do not show a significant difference in decoding performance between the OA and TA tasks. This suggests that, compared to deep learning models, traditional machine learning methods have certain limitations in capturing the underlying features of EEG signals. ML models rely on manual feature extraction and shallow learning strategies, making it difficult to fully exploit their potential in decoding complex auditory attention tasks, particularly when dealing with EEG signals that have high noise levels and variability. In contrast, deep learning models can automatically extract deep spatiotemporal features from raw data, which enhances their performance in decoding complex tasks like OA and TA task.

### B. Ablation Study

Batch Normalization (BN) ensures that the input values of non-linear transformation functions fall within a range sensitive to the inputs, thus preventing gradient vanishing [41], [42], [43]. To verify the effectiveness of BN in the AADNet model, we conducted a regularization ablation study. We examined the impact of ablating different modules on performance. Considering that we used BN layers in the temporal learning module, spatial learning module, and hybrid decoding module (BN1, BN2, BN3), we conducted three sets of ablation experiments on the proposed model using a 0.5s decision window as a baseline:

- M1: No BN1, with BN2, with BN3.
- M2: With BN1, no BN2, with BN3.
- M3: With BN1, with BN2, no BN3.

For the temporal learning module ablation (M1), the orientational task ACC decreased by 3% and the timbre task ACC decreased by 4%. For the spatial learning module ablation (M2), the orientational task ACC decreased by 2% and the timbre task ACC decreased by 3%. For the hybrid decoding module ablation (M3), the orientational task ACC decreased by 9% and the timbre task ACC decreased by 13%. Detailed results are shown in Table VIII.

| Task | | ACC | SPE | SEN | PRE | F1 |
|---|---|---|---|---|---|---|
| OA | M1 | 90.39 ± 4.47 | 89.54 ± 6.99 | 91.01 ± 6.71 | 90.77 ± 5.23 | 90.56 ± 4.59 |
| | M2 | 91.31 ± 4.13 | 90.31 ± 6.19 | 92.10 ± 5.48 | 91.33 ± 5.02 | 91.49 ± 4.17 |
| | M3 | 83.49 ± 6.84 | 82.82 ± 8.14 | 83.87 ± 8.00 | 84.32 ± 7.00 | 83.65 ± 7.13 |
| TA | M1 | 86.52 ± 4.97 | 83.72 ± 9.41 | 88.85 ± 5.46 | 84.37 ± 7.30 | 86.16 ± 4.89 |
| | M2 | 87.90 ± 4.84 | 85.33 ± 8.45 | 90.19 ± 4.49 | 85.80 ± 6.81 | 87.64 ± 4.67 |
| | M3 | 77.98 ± 6.74 | 78.58 ± 8.45 | 75.94 ± 11.15 | 77.04 ± 9.62 | 75.83 ± 10.06 |

## C. Data Distribution Visualization for AADNet

The original high-dimensional EEG data from the cue-masked auditory attention experiment, which are complex and not directly distinguishable, were visualized in two dimensions using PCA, as shown in Fig. 6 (A) and Fig. 6 (B). The data for male and female timbre and left and right orientational directions overlap. AADNet extracted 64-dimensional markers for orientation and timbre attention recognition.

For visualization purposes, these 64-dimensional markers were reduced to 2 dimensions using PCA, where each point represents a decision window, as depicted in Fig. 6 (C) and Fig. 6 (D). The timbre attentional markers and orientational attentional markers demonstrate good separability.

## IV. DISCUSSION

This study proposed a fully randomized paradigm for target orientation and timbre in a hybrid speech stimulus experiment, making it more suitable for real-time application scenarios. We collected EEG data from 16 subjects and introduced a portable, lightweight AADNet model for dual-task decoding. The proposed AADNet model possesses the capability to decode spatiotemporal features, effectively capturing information from shorter decision windows, and demonstrating excellent performance on the cue-masked auditory attention dataset.

## A. Low-Latency AADNet

From the comparisons across different decision window dimensions, it is evident that the proposed AADNet model outperforms other methods in both short (0.1s) and longer (1s) decision windows. In the 0.1s decision window, AADNet's ACC surpasses DL methods, specifically EEGNet and DeepCovNet, by 4% and 12% in OA decoding, and by 5% and 12% in TA decoding, respectively. Compared to the ML method PCA+SVM, AADNet is 36% higher in OA decoding and 31% higher in TA decoding. In the 0.5s decision window, AADNet's ACC in OA decoding exceeds EEGNet, ShallowCovNet, and DeepCovNet by 5%, 11%, and 12%, respectively, and outperforms FBCSP+SVM and PCA+SVM by 32% and 38%, respectively. In TA decoding, AADNet outperforms EEGNet, ShallowCovNet, and DeepCovNet by 6%, 14%, and 15%, respectively, and FBCSP+SVM and PCA+SVM by 35% and 33%, respectively. In the 1s decision window, AADNet exceeds the ACC of DL methods (EEGNet, ShallowCovNet, DeepCovNet) by over 10 percentage points and outperforms ML methods (FBCSP+SVM, PCA+SVM) by over 25 percentage points. In OA decoding, AADNet surpasses EEGNet, ShallowCovNet, and DeepCovNet by 10%, 13%, and 13%, respectively, and FBCSP+SVM and PCA+SVM by 33% and 34%, respectively. In TA decoding, AADNet surpasses EEGNet, ShallowCovNet, and DeepCovNet by 11%, 14%, and 13%, respectively, and FBCSP+SVM and PCA+SVM by 26%.

Aside from the proposed AADNet, among DL models, EEGNet performs well, achieving 7% higher accuracy than DeepCovNet in both OA and TA decoding in the 0.1s decision window. In the 0.5s decision window, EEGNet's OA decoding ACC is 6% and 7% higher than ShallowCovNet and DeepCovNet, respectively, and in TA decoding, it is 7% and 8% higher, respectively. In the 1s decision window, the performance differences among the other three DL models are minimal. The other two DL methods, ShallowCovNet and DeepCovNet, show comparable performance across different decision windows in both OA and TA decoding. Among ML models, FBCSP+SVM and PCA+SVM show similar performance across different decision windows and tasks. In OA and TA decoding, the mean ACC across different decision windows for DL methods (AADNet, EEGNet, ShallowCovNet, DeepCovNet) are 86.27% (0.1s), 86.14% (0.5s), and 79.64% (1s) in OA decoding, and 83.78%, 82.06%, and 74.12% in TA decoding. For ML methods (FBCSP+SVM, PCA+SVM), the mean ACC in OA decoding across different decision windows are 55.78%, 58.25%, and 55.26%, and in TA decoding are 58.26%, 56.72%, and 57.48%. In OA decoding, DL methods outperform ML methods by an average of 30% (0.1s), 27% (0.5s), and 24% (1s). In TA decoding, DL methods outperform ML methods by an average of 25% (0.1s), 25% (0.5s), and 16% (1s). In summary, DL methods outperform ML methods in both OA and TA decoding from 0.1s to 1s decision windows.

As expected, our method not only captures more fine-grained features but also exhibits time-insensitive characteristics. It is evident that the proposed AADNet model extracts sufficient EEG spatiotemporal information within the short decision window (0.1s), demonstrating superior and stable performance, with no significant improvement in ACC with increased decision window length. In the 0.1s decision window, the proposed AADNet model outperforms other models in both OA and TA decoding. In OA decoding, AADNet's average ACC is 8% higher than DL methods and 36% higher than ML methods; in TA decoding, AADNet's average ACC is 9% higher than DL methods and 31% higher than ML methods.

The low-latency design of AADNet ensures effective decoding within 0.1-second windows, making it suitable for real-time auditory attention applications. This rapid response capability is crucial for systems requiring fast and reliable attention decoding, such as brain-computer interfaces and real-time cognitive monitoring systems, where delays longer than 100 ms can hinder responsiveness. The model's architecture is optimized to capture essential spatiotemporal information swiftly, enabling it to maintain high accuracy even within short time windows, a distinctive advantage over other DL and ML models.

### B. Attention Guidance rather than Stimulation

In the latest ASAD dataset, subjects are informed in advance about the orientation of the sound they need to attend to, while sounds from other directions are designated as noise to be ignored. This paradigm might lead subjects to pre-allocate their attention to the indicated orientation, which is inconsistent with real-life scenarios where individuals receive and attend to sound signals spontaneously within short timeframes. This discrepancy may result in unobjective outcomes. To address this issue, we innovatively propose the most advanced cue-masked auditory attention paradigm, a cue-obscured mixed auditory attention framework. This paradigm obfuscates subjects' directional focus of attention by requiring them to selectively focus on the content of the target timbre speech segments based on timbre cues and to locate the position of the audio output. Concurrently, the content and location of non-target timbre speech segments are to be ignored as noise interference. By using target male and female voice cues to obscure the source location hints, our proposed paradigm is more complex and realistic. We aim to capture the shift of attention from an inattentive to an attentive state, which is more reflective of the natural attention process than pre-allocated attention. Our analysis suggests that timbre attention and orientational attention are decodable at different dimensional granularity.

Additionally, data forms the foundation for research on online neuro-steered hearing devices. While studies using public datasets are valuable, they have limitations in supporting ongoing research. Although our current study is limited to offline datasets, proposing an excellent paradigm, achieving data collection, and discovering reliable methods are the initial steps towards developing an online system.

### C. Orientation Prioritized, High Frequency Sensitized

For different decoding tasks, OA decoding and TA decoding, with DL methods in 0.1s, 0.5s, and 1s decision windows, OA decoding's ACC and SPE are higher than TA decoding. For AADNet, OA decoding's ACC is 1%, 2%, and 5% higher than TA decoding (for 0.1s, 0.5s, and 1s decision windows), and OA decoding's SPE is 5%, 4%, and 8% higher than TA decoding. For EEGNet, OA decoding's ACC is 3%, 3%, and 6% higher than TA decoding, and OA decoding's SPE is 5%, 4%, and 5% higher. For DeepCovNet, OA decoding's ACC is 2%, 4%, and 4% higher than TA decoding, and OA decoding's SPE is 4%, 11%, and 10% higher. For

ShallowCovNet, in the 0.5s and 1s decision windows, OA decoding's ACC is 5% and 6% higher than TA decoding, and OA decoding's SPE is 6% and 9% higher. Analyzing the methods and tasks together, the reason is that different timbres appear in different orientations, leading to higher ACC and SPE for OA decoding compared to TA decoding. Notably, in the proposed model, OA decoding's SPE is significantly higher than TA decoding's SPE. Additionally, there are instances where TA decoding's SEN is higher than OA decoding's SEN. In the AADNet method, in the 0.1s decision window, TA decoding's SEN is 1% higher than OA decoding. In the DeepCovNet method, in the 0.5s decision window, TA decoding's SEN is 0.3% higher than OA decoding.

Based on the experimental results (Table IV and Table V), we infer that the subjects' attention to the left sound source is superior to their attention to low-frequency sounds (male voices). This may be due to the auditory neural encoding mechanism, where the right hemisphere processes sound and spatial attention tasks more efficiently [44], [45]. Additionally, OA decoding consistently outperformed across various metrics. Our study provides evidence from another dimension that the human brain is more sensitive to high-frequency sound signals than to low-frequency signals, consistent with previous research conclusions [46]. Most existing papers primarily focus on OA decoding [15], [16], [17], with few comparisons between OA and TA. Our experimental results suggest that OA may be more suitable for decoding in AAD than TA, potentially due to the brain's preference for processing spatial localization.

### D. Limitation and Future work

The current experiment involved 16 participants. The dataset will be expanded in the future, with the long-term goal of applying the cue-masked auditory attention experimental paradigm to collect auditory attention EEG data from actual hearing-impaired individuals. Besides, though a multi-task cue-masked auditory attention paradigm has been proposed, this study focused on single-task decoding and decision-making. Multi-task decoding would obtain more properties of the sound source, thus achieving precise source identification. AADNet could be employed for the multi-task decoding scenario by adapting the classifier section. Currently, the prioritization of timbre and orientation processing in the brain is based on inferred outcomes, and further verification to explore the complex neural mechanisms of auditory attention requires more relevant experiments. Another limitation was that this study employed sample-based cross validation, which might improve the classification performance of the model. However, the comparison was fair for the performance of all the methods were evaluated under the same data splitting method and the same dataset. Trial-based cross validation was close to the scenario of real-world application, and should be adopted in future studies. The performance of the model on detecting sound orientation and timbre simultaneously would be studied, and the channel optimization would be conducted as well for precise auditory attention with a light and portable device, facilitating its application in daily life.

## V. Conclusion

This study proposed a novel cue-masked auditory attention experimental paradigm and an end-to-end learning model for fast and accurate auditory attention detection from EEG. The proposed paradigm overcomes the limitations of existing paradigms by authentically simulating the auditory attention focusing process. The proposed model effectively integrates spatial and temporal domains, capturing more fine-grained features of EEG signals. It significantly outperforms the current state-of-the-art models in both orientational and timbre decoding, and offers a higher level of interpretability in EEG signal decoding, independently addressing orientational and timbre decoding as separate dimensions. This work achieved high auditory attention decoding performance with low latency (>90% in 0.5 s) and without voice source knowledge, benefiting the potential applications of neuro-steered hearing aids and other assistive listening devices in real-world scenarios.

## VI. Acknowledgments

We thank the study participants for their involvement in this research, which made this discovery possible, and for granting permission to publish the information depicted in Figure 1. We thank Zhenyang Qin for assisting with the setup of the experiment. We thank Heshan Wang for algorithm design discussion. We thank Vivian Li for helping with figure editing and data collection.